\documentclass[conference]{IEEEtran}
\IEEEoverridecommandlockouts
\usepackage{cite}
\usepackage{algorithmic}
\usepackage{textcomp}
\usepackage{xcolor}
\def\BibTeX{{\rm B\kern-.05em{\sc i\kern-.025em b}\kern-.08em
    T\kern-.1667em\lower.7ex\hbox{E}\kern-.125emX}}

\usepackage{fancyhdr}
\usepackage{subcaption}
\usepackage{amsmath}
\usepackage{amssymb}
\usepackage{booktabs}
\usepackage{soul} 
\usepackage{amsthm}
\usepackage{amsfonts}
\usepackage{stmaryrd}
\usepackage{dsfont}
\usepackage{xfrac}
\usepackage{mathrsfs}
\usepackage{dsfont}
\usepackage{multirow}
\usepackage{tabularx}
\usepackage{tabulary}
\usepackage{tcolorbox}
\usepackage{colortbl}
\tcbuselibrary{most}
\usepackage{setspace}
\usepackage[bb=dsserif]{mathalpha}
\usepackage{makecell}
\usepackage{float}
\usepackage{amsmath}
\usepackage{ifthen}
\usepackage{pifont}

\usepackage{graphicx}
\usepackage{svg}
\usepackage{adjustbox}
\usepackage{longtable}
\usepackage{ulem} 
\newcommand{\cmark}{\textcolor{black}{\ding{51}}}
\newcommand{\xmark}{\textcolor{black}{\ding{55}}}

\definecolor{darkred}{RGB}{139, 0, 0}
\definecolor{darkblue}{RGB}{0, 0, 139}
\definecolor{darkgreen}{RGB}{0, 200, 0} 
\definecolor{ocra}{RGB}{204, 119, 34}

\newcommand{\ours}{FreeSeg-Diff}

\definecolor{cvprblue}{rgb}{0.21,0.49,0.74}
\usepackage[pagebackref,breaklinks,colorlinks,citecolor=cvprblue]{hyperref}
\usepackage[capitalize]{cleveref} 

\begin{document}

\title{\ours: Training-Free Open-Vocabulary Segmentation with Diffusion Models}  


\author{
\vspace{1em}
\begin{tabular}{ccc}
\begin{tabular}[t]{c}
Barbara Toniella Corradini \\
\textit{DIISM} \\
University of Siena \\
Siena, Italy \\
barbara.corradini@unisi.it
\end{tabular}
&
\begin{tabular}[t]{c}
Mustafa Shukor \\
\textit{CNRS, ISIR} \\
Sorbonne University \\
F-75005 Paris, France \\
mustafa.shukor@isir.upmc.fr
\end{tabular}
&
\begin{tabular}[t]{c}
Paul Couairon\\
\textit{CNRS, ISIR} \\
Sorbonne University \\
F-75005 Paris, France \\
paul.couairon@isir.upmc.fr
\end{tabular}
\\[2.2cm]
\begin{tabular}[t]{c}
Guillaume Couairon \\
\textit{Inria, ARCHES} \\
Sorbonne University \\
F-75005 Paris, France \\
guillaume.couairon@gmail.com
\end{tabular}
&
\begin{tabular}[t]{c}
Franco Scarselli \\
\textit{DIISM} \\
University of Siena \\
Siena, Italy \\
franco.scarselli@unisi.it
\end{tabular}
&
\begin{tabular}[t]{c}
Matthieu Cord \\
\textit{CNRS, ISIR; Valeo.ai} \\
Sorbonne University \\
F-75005 Paris, France \\
matthieu.cord@sorbonne-universite.fr
\end{tabular}
\end{tabular}
}

\twocolumn[{%
\renewcommand\twocolumn[1][]{#1}%
\maketitle

\begin{center}
    \centering
    \captionsetup{type=figure}
    \vspace{-1cm}
    \includegraphics[width=0.63\textwidth]{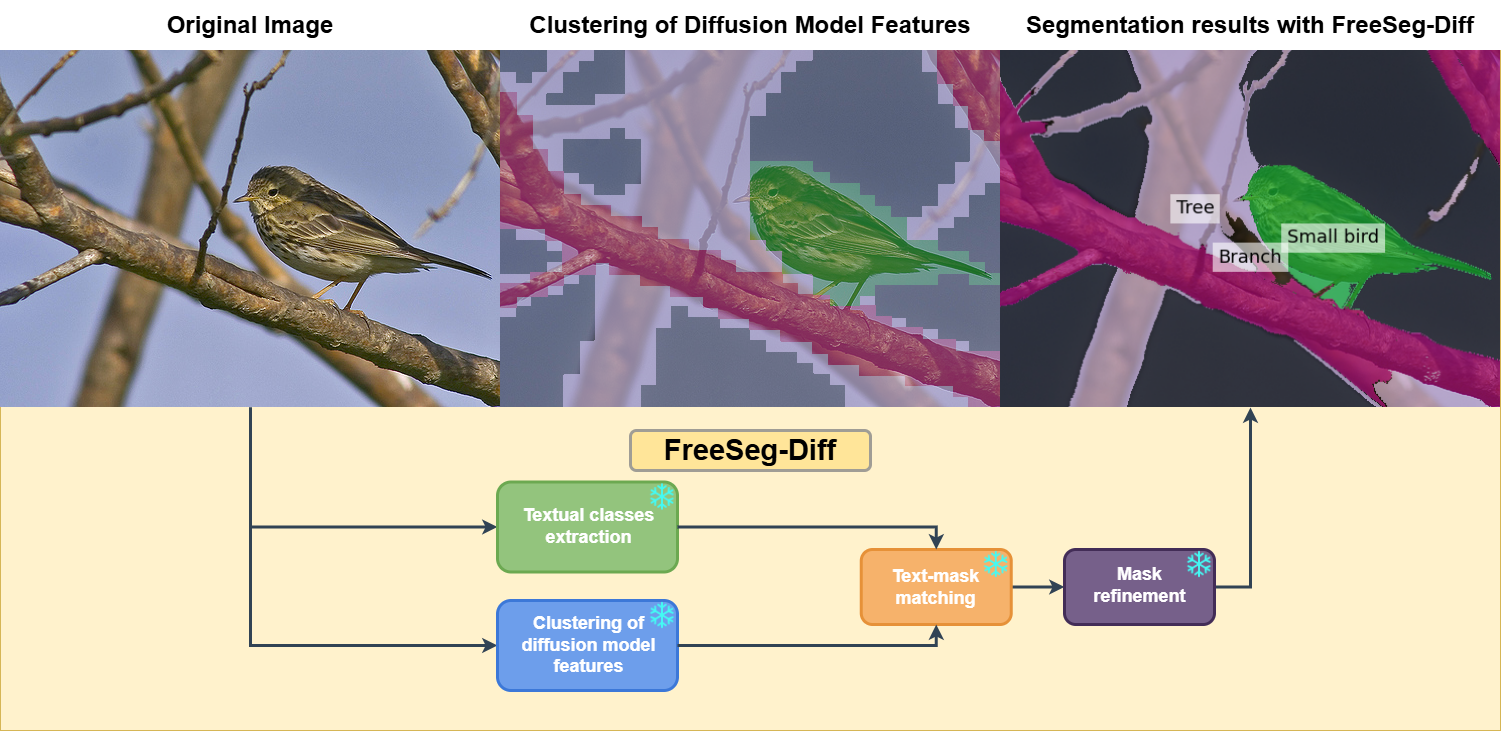}
    \caption{
    \textbf{Zero-shot open-vocabulary segmentation with \ours{}}. 
    The image is passed to a diffusion model and to an image captioner model to get visual features and textual description, respectively. The features are used to get class-agnostic masks, which are then associated with the extracted text. The final segmentation map is obtained after a mask refinement step. All models are kept frozen.}
    \label{fig:main}
 \end{center}%
}]

\begin{abstract}
Foundation models have exhibited unprecedented capabilities across various domains and tasks. 
Models like CLIP bridge cross-modal representations, while text-to-image diffusion models excel in realistic image generation.
While the complexity of these models makes retraining infeasible, their superior performance has driven research to explore how to efficiently use them for downstream tasks. 
Our work explores how to leverage these models for dense visual prediction tasks, specifically image segmentation.
To avoid the annotation cost or training large diffusion models, we constrain our method to be zero-shot and training-free. 
Our pipeline, dubbed FreeSeg-Diff, uses open-source foundation models to perform open-vocabulary segmentation as follows: (a) retrieving image caption (via BLIP-2) and visual features (via Stable Diffusion), (b) clustering and binarizing features to form class-agnostic object masks, (c) mapping these masks to textual classes using CLIP with open vocabulary support, and (d) refining coarse masks.
\ours{} surpasses many training-based methods on Pascal VOC and COCO datasets and delivers competitive results against recent weakly-supervised segmentation approaches. 
We provide experiments demonstrating the superiority of diffusion model features over other pre-trained models. 
Project page: \href{https://bcorrad.github.io/freesegdiff/}{https://bcorrad.github.io/freesegdiff/}.
\end{abstract}


\section{Introduction}
\label{sec:intro}
Foundation models~\cite{bommasani2021opportunities, zhou2023comprehensive, awais2023foundational}, \textit{i.e.}, large language models (LLMs) ~\cite{achiam2023gpt,team2023gemini,touvron2023llama} and multimodal models~\cite{radford2021learning, alayrac2022flamingo,shukor2023unified} are saturating across most of the benchmarks. 
Among these foundation models, generative models~\cite{ramesh2022hierarchical, brown2020language, touvron2023llama} have got most of the attention due to their impressive capabilities in text and image generation.

Image generative models~\cite{kang2023scaling, zhang2023text}, particularly diffusion-based models (DMs)~\cite{denoisingdiffusionprobabilisticmodels, denoisingdiffusionimplicitmodels, rombach2022high} have achieved unprecedented performance in generating images indistinguishable from real ones. 
Besides image quality, these models are easily controllable via different multimodal inputs, such as segmentation maps and text~\cite{ramesh2022hierarchical,zhang2023adding,couairon2023zero}. 
Despite significant efforts to develop more powerful generative foundation models~\cite{podell2023sdxl} or training larger models on extensive datasets, researchers are exploring how pre-trained foundation models can be used to learn downstream tasks through efficient fine-tuning~\cite{vallaeys2024improveddepalm, lialin2023scaling} or few-shot in-context learning~\cite{brown2020language, alayrac2022flamingo}.
Image segmentation~\cite{csurka2022semantic}, which involves pixel-wise classification, is one of the fundamental tasks in computer vision that drives a myriad of applications in many domains, such as medical imaging~\cite{Rahman_2023_CVPR}, robotics~\cite{tzelepi2021semantic,pan2020cross} and autonomous driving ~\cite{chen2023edge}.
This task is traditionally tackled by end-to-end training of deep learning models on densely annotated datasets. 
However, annotating each pixel in the image is a very tedious and costly task, which hinders the development of foundation models for segmentation. 
Several alternative paradigms have shown promise to tackle this task, such as semi-supervised~\cite{wang2022semi}, weakly-supervised~\cite{weakly} and unsupervised learning~\cite{hwang2019segsort}.
Another limitation of traditional supervised approaches is the closed-set vocabulary; models cannot segment novel objects and classes that are not in the training set. 
In the real world, models should be able to segment any object in the wild, which is often referred to as open-vocabulary segmentation~\cite{liang2023open, xu2022simple}.
While many efforts focus on adapting foundation models to tasks closely related to their original purpose (\textit{e.g.}, generative models adapted to edit images), here we go a step further and wonder:
\begin{center}
    \textit{How can we 
    leverage off-the-shelf image generative foundation models,
    for open-vocabulary segmentation?} 
\end{center}
A positive answer to this question could pave the way to exploit future generative models for downstream tasks while avoiding the cost associated with training and annotations. 
Inspired by works showing some form of localized information inside DMs~\cite{xu2023open, baranchuk2021label}, we try to answer by extracting the internal features of text-to-image diffusion models from an image, to find the regions corresponding to different objects (\Cref{fig:main}). 
Our contributions can be summarized as follows:
\begin{itemize}
    \item We propose \ours{}, a zero-shot training-free image segmentation pipeline using DMs, not relying on annotated masks or inference-time optimization.
    \item We provide a comparison between the features of DMs and other pre-trained models, showing the superiority of DMs features for localization.
    \item \ours{} outperforms many previous approaches that rely on training or optimization and compete with recent weakly-supervised SoTA, on both COCO and Pascal VOC datasets. 
\end{itemize}

\section{Related Works}
\label{sec:background}

\subsection{Diffusion Models}

Diffusion models have seen remarkable success in text-to-image generative applications such as DALL-E~\cite{ramesh2022hierarchical}, Imagen~\cite{wang2023imagen} or Stable Diffusion~\cite{rombach2022high} trained on Internet-scale data such as LAION-5B~\cite{schuhmann2022laion5b}. These approaches are driven by probabilistic generative models  ~\cite{denoisingdiffusionprobabilisticmodels, denoisingdiffusionimplicitmodels} that are trained to invert a diffusion process. 
In the forward diffusion process, the input image $x_0$ is gradually destroyed by the addition of Gaussian noise for a certain number of timesteps $T$. In the reverse diffusion process, a neural network is trained to predict the amount of noise which was added to the image at timestep $t$, namely $\epsilon_\theta(x_t, t)$, by minimizing the loss function with respect to the target added noise $\bar{\epsilon}$: $\mathcal{L}=\mathbb{E}_{x_0,t,\bar{\epsilon}}\parallel\bar{\epsilon} - \epsilon_\theta(x_t, t)\parallel ^2 _2$.
DMs have demonstrated superior performance compared to Generative Adversarial Networks (GANs)~\cite{goodfellow2014generative} in terms of the quality of generated images, while also mitigating limitations associated with GANs, such as mode collapse. 
However, the significant computational power required by these models has spurred new research directions aimed at enhancing their efficiency. 
To address the challenge of image generation within a high-dimensional pixel space, Stable Diffusion Models (SDMs)~\cite{rombach2022high} have been introduced. 
Unlike DMs, they encode the image in a lower-dimensional latent space and exploit cross-modal attention mechanism, allowing them to concentrate on the semantic features of the image and achieve greater computational efficiency. 
\subsection{Open-Vocabulary Zero-Shot Segmentation}
Zero-shot segmentation consists of segmenting using pre-trained models, without explicitly training on the target dataset~\cite{NEURIPS2019_0266e33d}. 
Open-vocabulary segmentation, \textit{i.e.}, segmenting any object in the wild without relying on a predefined set of classes, is a complex task that is currently tackled mainly using CLIP. 
MaskCLIP~\cite{dong2022maskclip} and CLIPSeg~\cite{Luddecke_2022} extend CLIP by integrating image and text inputs with an additional segmentation network to enhance segmentation tasks.
In~\cite{yu2023convolutions}, CLIP is recommended as a feature extractor for learning segmentation masks, owing to its superior generalization properties and performance when scaling up input image resolution compared to ViT-based backbones.
Similar to our work, ReCo~\cite{shin2022reco} generates segmentation masks using a caption that describes the content of an image. 
CLIP is employed for image retrieval by collecting exemplar images from ImageNet~\cite{imagenet} based on a given language query.
Recently, many open-vocabulary segmentation approaches~\cite{xu2022groupvit, ren2023viewco, xu2023learningovs} are based on weakly supervised approaches, where models are trained to segment objects based only on language supervision. 
However, most of these approaches rely on additional training and different kinds of supervision.

\subsection{Leveraging diffusion models for visual tasks} 
The success of DMs attracted a lot of effort to leverage such models beyond image generation. 
These models are currently used to tackle classical visual tasks, such as image classification~\cite{chen2023robustclassif, mukhopadhyay2023diffusionclassif}, object detection~\cite{chen2023diffusiondetect, zhang2023diffusionenginedetect}, and image segmentation. 
In particular, DiffuMask~\cite{wu2023diffumask} exploits cross-modal attention maps between image features and conditioning text embeddings to segment the most prominent object, referred to by a text prompt. DiffusionSeg~\cite{ma2023diffusionseg} first generates a synthetic image/mask dataset and then leverages DDIM inversion to obtain feature maps and attention masks of object-centric images to perform unsupervised object discovery. 
However, the method relies on ImageNet labels and is not open-vocabulary. 
LD-ZNet~\cite{pnvr2023ldznet} shows that the internal features of LDMs contain rich semantic information to perform text-based segmentation of synthetic images. 
In~\cite{baranchuk2021label} image segmentation is performed by training multi-layer perceptrons to classify pixels based on U-Net feature maps. Open-vocabulary segmentation is performed in~\cite{xu2023open} by a text-to-image diffusion model. 
To eliminate the need for dependence on a pre-trained image captioner, this work introduces the concept of an implicit image captioner. 
In~\cite{karazija2023ovdiff}, the authors tackle this challenge with a DM by identifying candidate classes and generating a set of supporting images, which are then utilized to identify insignificant portions of the image.
In~\cite{burgert2022peekaboo}, the authors introduce Peekaboo, which optimizes a parameter that generates an optimal mask for the textual prompt used as conditioning for the SDM. 

\begin{figure*}[t]
\centering
    \label{fig:pipeline_overview}
    \includegraphics[width=.83\textwidth]{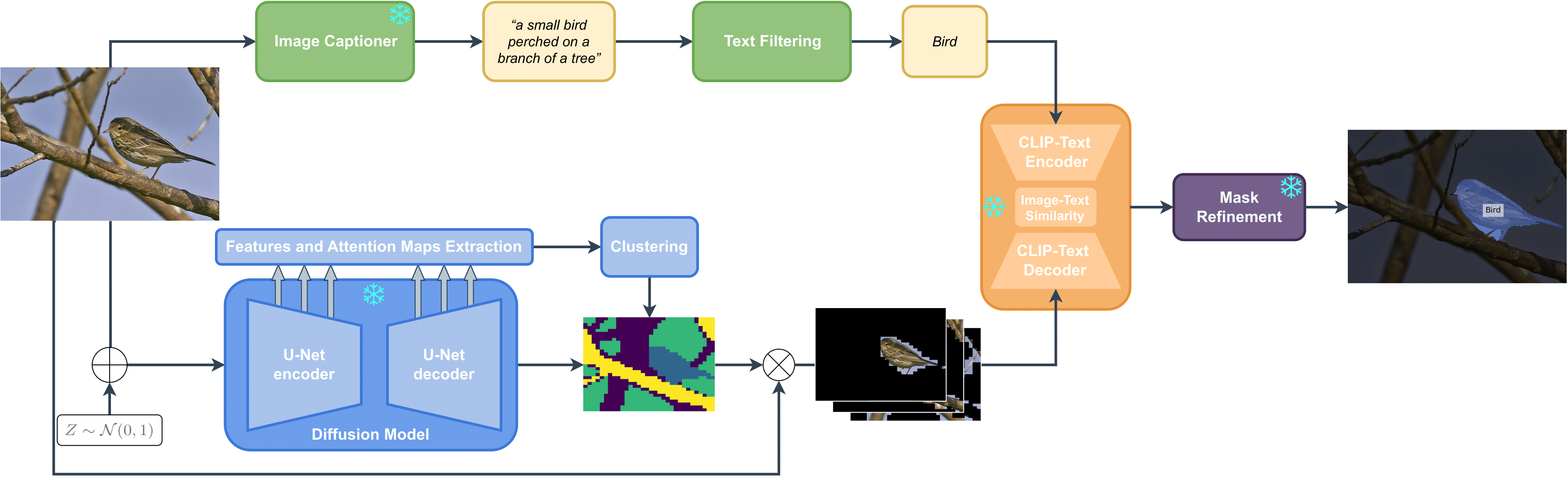}
    \caption{\textbf{Our detailed pipeline for zero-shot semantic segmentation.} 
    Image features extracted from the DM blocks are clustered to create class-agnostic masks, which are then superimposed on the original image and processed by CLIP to link each mask to a specific text (\textit{i.e.}, class). 
    Textual classes are derived from entities extracted from a BLIP caption. 
    After matching masks to classes, more accurate masks are obtained by a mask refinement module. 
    All models remain frozen.
    }
\end{figure*} 

\section{\ours{}} \label{sec:our_pipeline}
In this work, we introduce a pipeline to achieve object segmentation in a training-free zero-shot manner. An overview of our method is shown in~\Cref{fig:pipeline_overview}: 
the image is fed to a pre-trained caption generator (to get a textual description) and an SDM to extract the visual features.
In a closed-vocabulary setup, the caption is filtered to identify keywords associated with the predefined list of classes. 
To localize the objects in the image, we utilize an unsupervised clustering algorithm to obtain a set of class-agnostic masks from the extracted internal features.
Each mask is then superimposed on the image and associated with the text that most closely matches that portion of the image.
Finally, the masks are refined to obtain more accurate segmentation masks. 
We provide a more in-depth explanation of \ours{} in the following.
\subsubsection{Candidate classes assessment.}
\label{sec:class_filtering}
As semantic segmentation becomes more challenging with an increasing number of classes, we narrow down the list of potential classes to generate a list of candidate objects that might be present in the image. 
To this end, we query an off-the-shelf automatic image captioner~\cite{li2023blip2} to obtain a textual description of the visual content (\textit{e.g.}, \emph{A small bird perched on a branch of a tree}). 
We then use basic Natural Language Processing (NLP) methods to extract entities from the caption (\textit{i.e.}, \emph{bird}). 
Since, in a closed-vocabulary setup, entities may not match the dataset classes exactly (\textit{e.g.}, synonyms like \emph{couch} and \emph{sofa}), we choose the candidate classes that minimize the distance with the keywords in the latent space of a CLIP text encoder.
If the distance is above the average distances (to all classes), the keyword is ignored and not added to the candidate list. 
Note that using CLIP here is only to evaluate closed-vocabulary benchmarks, while it is not required for open-vocabulary segmentation tasks.

\subsubsection{Class-agnostic masks retrieval.}
\label{subsubsec:agnostic_mask_retrieval}
The U-Net backbone of an SDM is constructed as a succession of convolution and attention blocks, alternating with either upsampling or downsampling blocks, containing dense and information-rich representation, particularly beneficial for panoptic segmentation tasks~\cite{xu2023open}. 
The image is passed through the SDM for a single denoising step to get the visual features.
Instead of considering only the last layer of the DM, we extract multiscale features (\textit{i.e.}, encoder and decoder's resblock). 
Tensors are concatenated and clustered using the K-means algorithm. 

\subsubsection{Masks classification and refinement.}
While the clustering algorithm succeeds in dividing the internal representation of the SDM into clusters, the resulting coarse mask remains class-agnostic.
To associate pixels with one of the candidate classes, we mask the original image with the binarized version of each cluster. 
The masked image is encoded using CLIP, and the region is labelled according to the minimal embedding distance to the text. 
If the nearest class is in the candidate classes, the image region is assigned that class; otherwise, the region is ``unlabeled''.
After this phase, we get a coarse semantic segmentation map in which every pixel has been labelled. 
Finally, for each classified region of the image, the corresponding map is processed by Conditional Random Fields (CRF)~\cite{krahenbuhl2011efficient} to obtain the final refined semantic map.

\section{Experiments}
\label{sec:exp}
In this section, we report the implementation details of \ours{}, comparing our results against many SoTA methods. 
\subsection{Implementation Details}
\label{sec:implementation_details} 
To retrieve image visual features, we use Stable Diffusion~\cite{rombach2022high} v1-5 pre-trained on a subset of the LAION-5B~\cite{schuhmann2022laion5b} dataset. 
We set the time step used for the diffusion process to $t=0$ by default to encode the image without noise. 
We resize the internal representations extracted from each resblock of the U-Net to $32 \times 32$. 
We use CLIP VIT-B/32~\cite{radford2021learning} as our text-image discriminative model. 
Based on ablations on varying number of clusters $K$, the results for $K=4$ and $K=5$ are comparable, so we use $K=4$ for computational efficiency.
After feature clustering, we rescale the coarse mask to the original image dimensions with nearest neighbour interpolation. 
Finally, we use CRF algorithm to refine masks and obtain the final segmentation.
Following previous works, we evaluate the mIoU of \ours on the validation split of two popular benchmarks, Pascal VOC with 21 categories (20 classes + ``unlabeled'')~\cite{Everingham15} and COCO (COCO-Objects)~\cite{lin2014microsoft} with 81 categories (80 classes + ``unlabeled''). 
We also report results on the VOC-C dataset used in~\cite{burgert2022peekaboo}, a filtered part of Pascal VOC, and COCO-27, which is COCO with 27 supercategories instead of 80 classes. Further data and analysis are provided in the Supplementary Material.

\begin{table*}[t]
\centering
\caption{\textbf{Evaluation of semantic segmentation using \ours{}} on the Pascal VOC, VOC-C~\cite{burgert2022peekaboo}, COCO, and COCO-27 datasets. 
The results are based on K-Means clustering applied to SDM feature maps (at a resolution of $16$) with a fixed $K=4$. 
Models that do not require any training are highlighted in yellow. 
Deltas between our results and the proposed method are indicated in \textcolor{darkgreen}{green} and \textcolor{red}{red}. Results from: 1~\cite{xu2022groupvit}, 2~\cite{ranasinghe2022perceptual}, 3~\cite{cha2023learningtcl}.}
\label{tab:main_pascal}
\resizebox{0.75\linewidth}{!}{
  \begin{tabular}{lcccccc}
    \hline
    \textbf{} & 
    \multirow{2}{*}{\textbf{DMs}} &
    \multirow{2}{*}{\thead{\textbf{Supervision}}}  & 
    \multicolumn{4}{c}{\thead{\textbf{mIoU}}} \\
    & & & Pascal VOC & VOC-C & COCO & COCO-27 \\
    \hline
    MoCo (IN)  $^1$ & \xmark & Masks &  34.3 \textcolor{darkgreen}{(+18.97)} & -- & -- & -- \\
    DINO (IN)  $^1$ & \xmark & Masks &  39.1 \textcolor{darkgreen}{(+14.17)} & -- & -- & -- \\
    DeiT (IN)  $^1$ & \xmark & Masks &  53.0 \textcolor{darkgreen}{(+0.27)} & -- & -- & -- \\
    \midrule
    \rowcolor{yellow!20} ReCo $^3$ & \xmark & Text &  25.1 \textcolor{darkgreen}{(+28.17)} & -- & 15.7 \textcolor{darkgreen}{(+15.31)} & 26.3 \textcolor{darkgreen}{(+7.73)} \\ 
    ViL-Seg   & \xmark & Text &  37.30 \textcolor{darkgreen}{(+15.97)} & -- & -- & -- \\
    ViewCo   & \xmark & Text & 45.7 \textcolor{darkgreen}{(+7.57)} & -- & 23.5 \textcolor{darkgreen}{(+7.51)} & -- \\ 
    OVSegmentor & \xmark & Text &  53.8 \textcolor{red}{(-0.53)} & -- & 25.1 \textcolor{darkgreen}{(+5.91)} & -- \\
    GroupViT & \xmark & Text &  52.3 \textcolor{darkgreen}{(+0.97)} & 57.8 \textcolor{darkgreen}{(+1.06)} & 24.3 \textcolor{darkgreen}{(+6.71)} & -- \\
    SegCLIP  & \xmark & Text &  52.6 \textcolor{darkgreen}{(+0.67)} & -- & 26.5 \textcolor{darkgreen}{(+4.51)} & -- \\
    TCL    & \xmark & Text &  55.0 \textcolor{red}{(-1.73)} & -- & 31.6 \textcolor{red}{(-0.59)} & -- \\ 
    CLIPpy (134M) & \xmark & Text &  52.2 \textcolor{darkgreen}{(+1.07)} & -- & 32.0 \textcolor{red}{(-0.99)} & -- \\
    \midrule
    \rowcolor{yellow!20} CLIP $^2$ & \xmark & Text & 18.1 \textcolor{darkgreen}{(+35.17)} & -- & 14.5 \textcolor{darkgreen}{(+16.51)} & -- \\
    \rowcolor{yellow!20} MaskCLIP $^2$& \xmark & Text & 22.1 \textcolor{darkgreen}{(+31.17)} & -- & 13.8 \textcolor{darkgreen}{(+17.21)} & 19.6 \textcolor{darkgreen}{(+14.43)} \\ 
    \rowcolor{yellow!20} K-Means + CLIP & \xmark & Text &  25.0 \textcolor{darkgreen}{(+28.27)} & -- & -- & -- \\
    \rowcolor{yellow!20} ALIGN $^2$ & \xmark & Text &  29.7 \textcolor{darkgreen}{(+23.57)} & -- & -- & -- \\
    \midrule
    \rowcolor{yellow!20} Peekaboo & \cmark & Text &  -- & 52.0 \textcolor{darkgreen}{(+6.86)} & -- & -- \\
    \rowcolor{yellow!20} OVDiff & \cmark & Text + Pseudo Masks & 69.0 \textcolor{red}{(-15.73)} & -- & 36.3 \textcolor{red}{(-5.29)} & -- \\ 
    DiffuMask & \cmark & Text + Pseudo Masks & 70.6 \textcolor{red}{(-17.33)} & -- & -- & -- \\ 
    ODISE & \cmark & Text + Masks &  82.7 \textcolor{red}{(-29.43)} & -- & 52.4 \textcolor{red}{(-21.39)} & -- \\  
    \midrule
    \rowcolor{yellow!20} \textbf{\ours{} (Ours)} & \cmark & Text & 53.27 & 58.86 & 31.01 & 34.03 \\
    \hline
  \end{tabular}
}
\end{table*}

\subsection{Comparison with State-of-The-Art Methods}
\label{sec:sota_comparison}

This section compares \ours{} to existing methods on popular image segmentation datasets, specifically Pascal VOC and COCO. In~\Cref{tab:main_pascal}, we present results categorized by algorithm type: mask-supervised, text weakly-supervised, text-supervised, and diffusion-based approaches. Each category specifies the type of supervision (Masks, Text, or Text + Pseudo Masks), with approaches that require no training highlighted in yellow. We explore the comparison with other segmentation approaches further below.

\subsubsection{Mask supervised approaches.} 
We compare our method with non-DM-based approaches that explicitly use target segmentation masks for supervision. 
These approaches utilize visual models pre-trained on ImageNet for image classification, including fully supervised methods like DeiT~\cite{touvron2021trainingdeit} and self-supervised methods such as DINO~\cite{caron2021emergingdino} and MoCo~\cite{he2020momentummoco}.
As shown in~\Cref{tab:main_pascal}, \ours{} (which exploits text-weak supervision instead of mask supervision) outperforms MoCo (+19 points) and DINO (+14 points).  

\subsubsection{Text weakly-supervised approaches.} 
Recently, many methods have been developed for segmenting images using only textual supervision, \textit{i.e.}, localizing objects from image captions. \ours{} outperforms these methods in a zero-shot setup on both Pascal VOC (+28 points over ReCo) and COCO (+15 over ReCo~\cite{shin2022reco}, +7 over ViewCo~\cite{ren2023viewco}). 
Notably, our model is also competitive with more recent advanced methods such as CLIPpy~\cite{ranasinghe2022perceptual}, OVSegmentor~\cite{xu2023learningovs}, and TCL~\cite{cha2023learningtcl}.

\subsubsection{Training-free text-based approaches.} 
We compare our method against approaches that do not use training for localization, such as clustering techniques (\textit{e.g.}, K-Means) and feature-based models (\textit{e.g.}, MaskCLIP~\cite{zhou2022extractmaskclip}). 
As shown in~\Cref{tab:main_pascal}, we significantly outperform these methods by up to 29 points on Pascal VOC, 15 points on COCO-27, and 16 points on COCO. 
These approaches, which often rely on global contrastive image-text alignment (\textit{e.g.}, CLIP~\cite{radford2021learning} and ALIGN~\cite{align}), generally lack the fine-grained, localized spatial information.

\subsubsection{Diffusion Model-based approaches.} 
Recent efforts aim to segment by using generative models, in particular, diffusion models.~\Cref{tab:main_pascal} shows that \ours{} outperforms   Peekaboo~\cite{burgert2022peekaboo}, which additionally relies on inference-time optimization, on Pascal VOC-C (+6 points). 
However, we still underperform models that leverage diffusion models and do additional training for segmentation (\textit{e.g.}, ODISE~\cite{xu2023open}). 
These results show the superiority of generative models (\textit{e.g}, compared to CLIP) when it comes to localization capability.

\subsection{Qualitative results}
We show some qualitative results of \ours{} in~\Cref{fig:qualitatives}. 
Compared to the raw clustering (row 2), the final masks are more refined and tightly cover the objects. 
Another advantage of our method is the implicit cluster filtering due to our text processing branch (BLIP + text entity extraction). 
Due to the general representation of CLIP, the model can associate well the class-agnostic masks with the correct label. 
\newcommand{\InclGr}[1]{\includegraphics[width=1.2cm,height=1.2cm]{#1}}
\begin{figure*}[t]
    \centering
    \begin{subfigure}[t]{0.3\linewidth}
        \centering
        \renewcommand{\arraystretch}{0}
        \begin{tabular}{*6{@{}c}}
            \InclGr{}&
            \InclGr{}&
            \InclGr{}&
            \InclGr{}&
            \\
            \InclGr{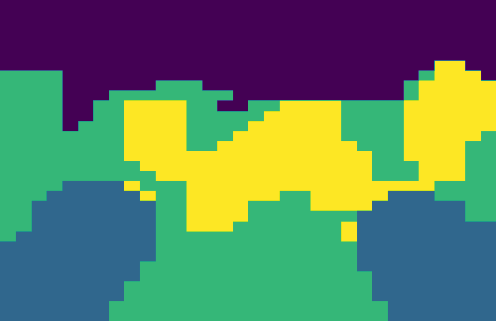}&
            \InclGr{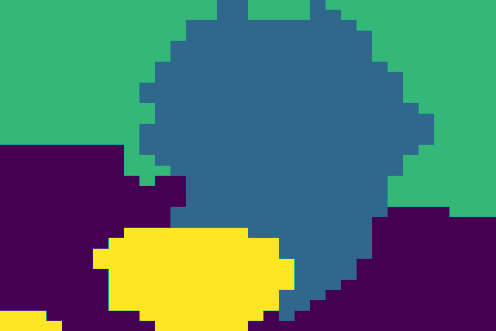}&
            \InclGr{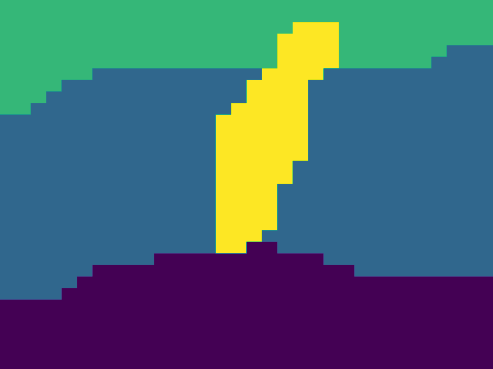}&
            \InclGr{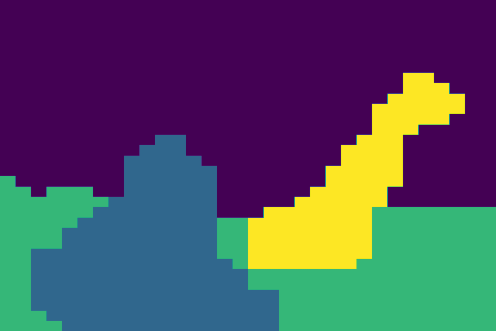}&
            \\
            \InclGr{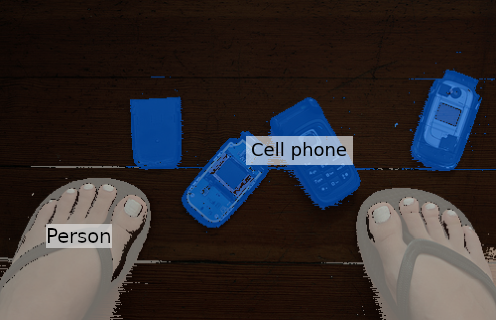}&
            \InclGr{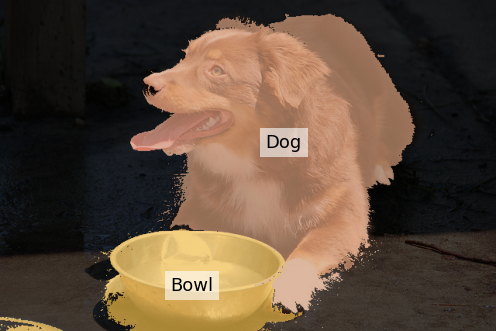}&
            \InclGr{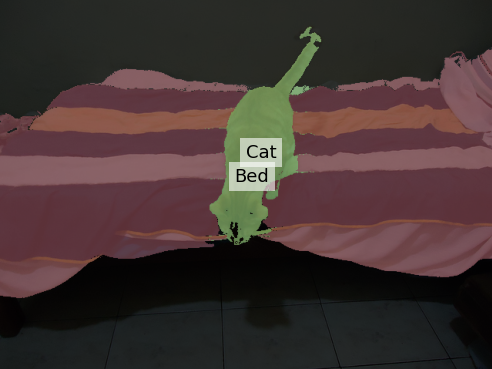}&
            \InclGr{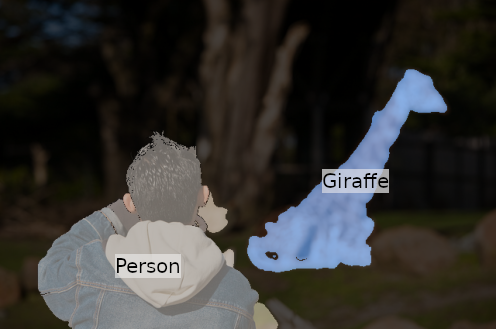}&
        \end{tabular}
    \caption{Closed-vocabulary segmentation on COCO.}
    \label{fig:coco_qual_results}
    \end{subfigure}
    \begin{subfigure}[t]{0.3\linewidth}
        \centering
        \renewcommand{\arraystretch}{0}
        \begin{tabular}{*6{@{}c}}
            \InclGr{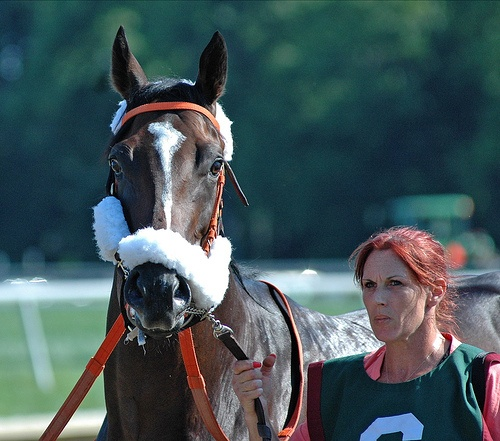}&
            \InclGr{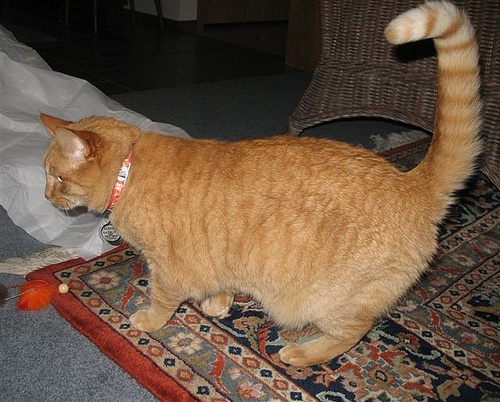}&
            \InclGr{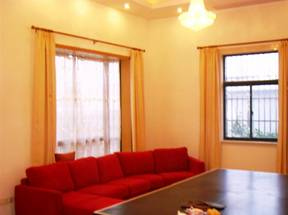}&
            \InclGr{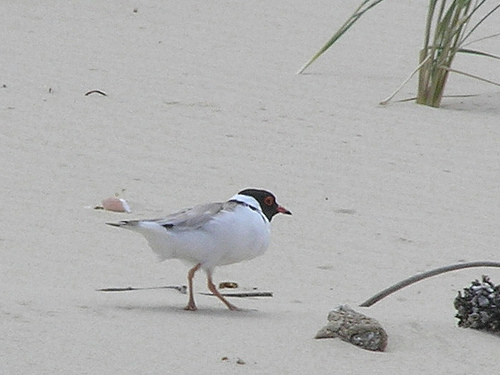}\\
            \InclGr{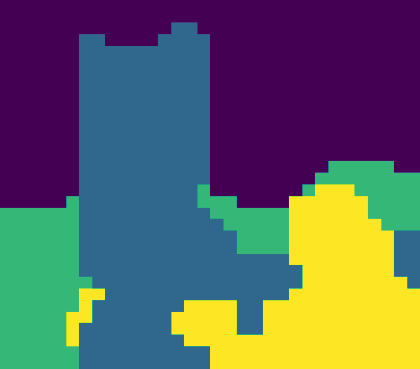}&
            \InclGr{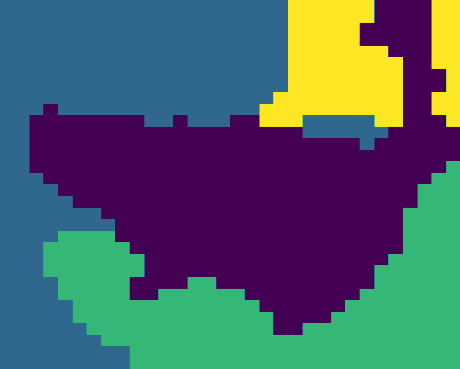}&
            \InclGr{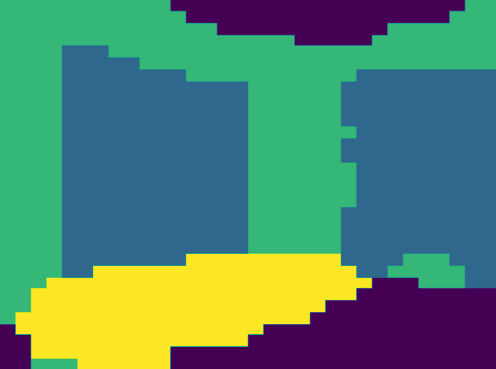}&
            \InclGr{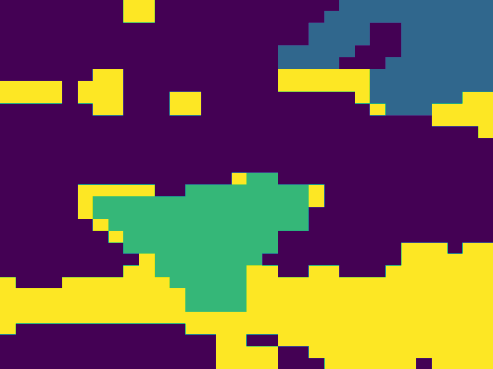}\\
            \InclGr{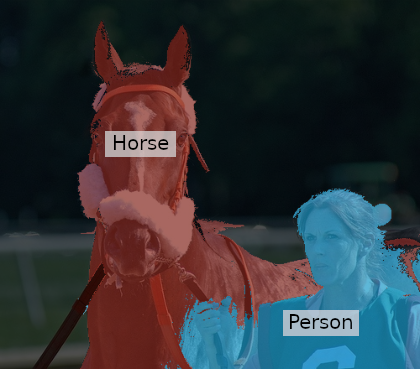}&
            \InclGr{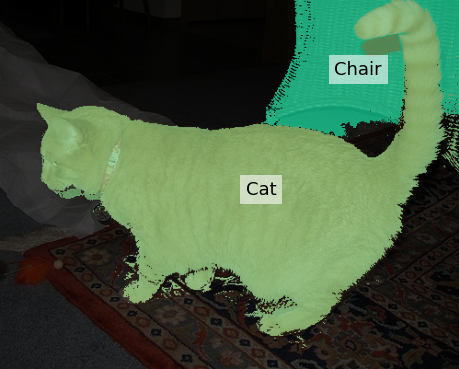}&
            \InclGr{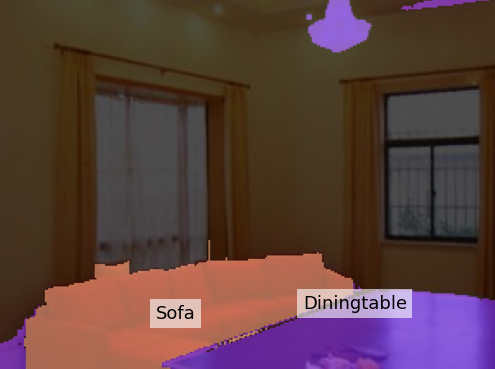}&
            \InclGr{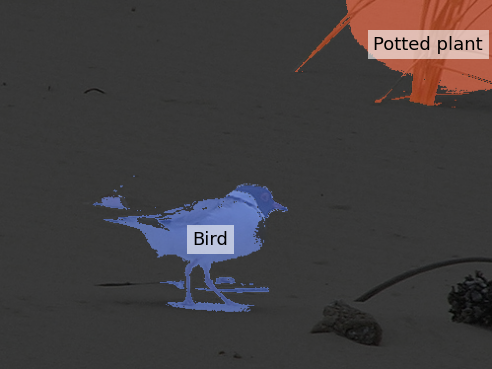}
        \end{tabular}
    \caption{Closed-vocabulary segmentation on Pascal VOC.}
    \label{fig:voc_multi_qual_results}
    \end{subfigure}
    \begin{subfigure}[t]{0.3\linewidth}
    \centering
    \renewcommand{\arraystretch}{0}
    \begin{tabular}{*4{@{}c}}
        \InclGr{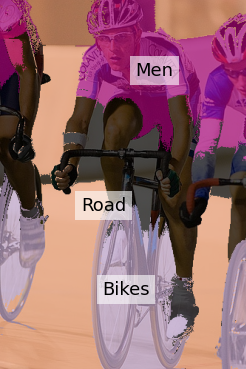}&
        \InclGr{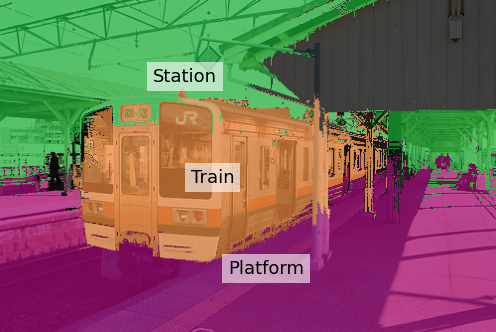}&
        \InclGr{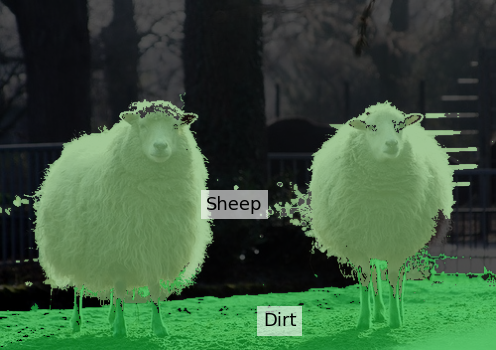}&
        \InclGr{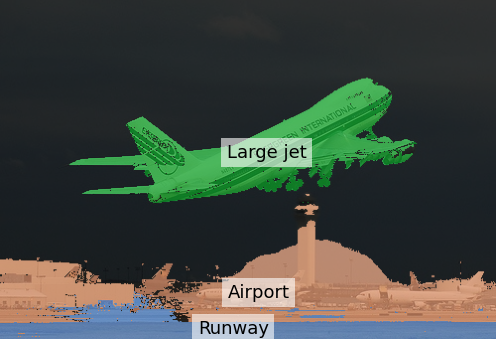}\\
        \InclGr{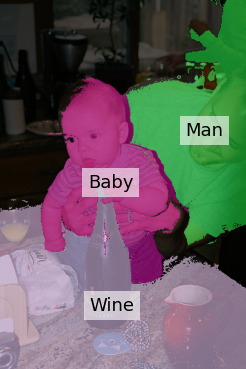}&
        \InclGr{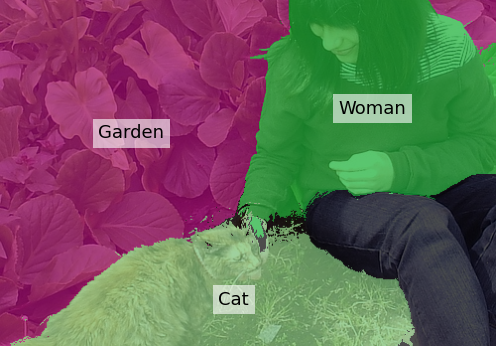}&
        \InclGr{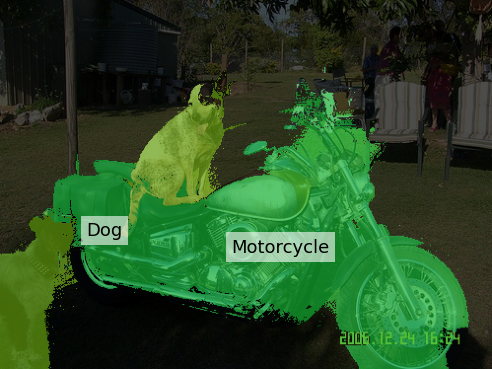}&
        \InclGr{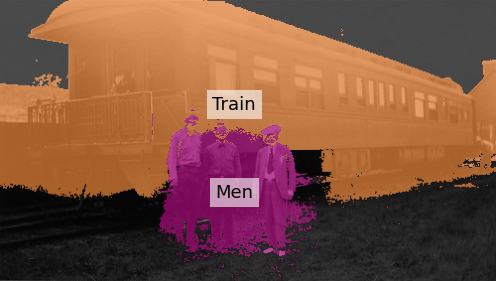}\\
        \InclGr{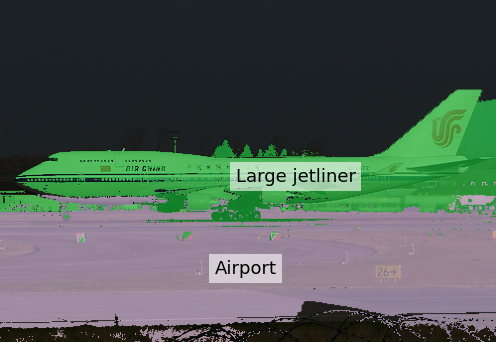}&
        \InclGr{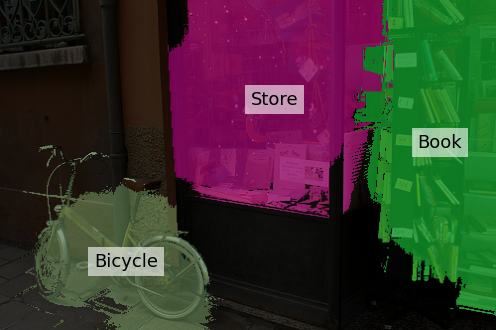}&
        \InclGr{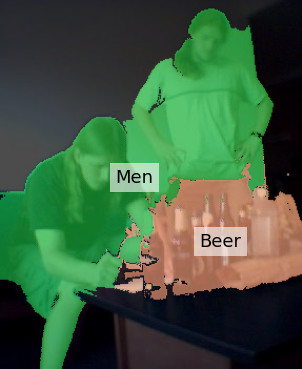}&
        \InclGr{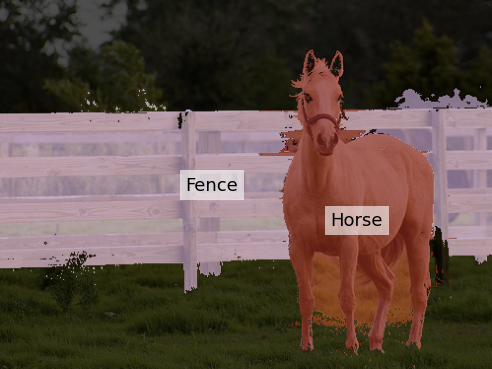}
    \end{tabular}    
    \caption{Open-vocabulary segmentation on Pascal VOC.}
    \label{fig:openvoc-qual}
    \end{subfigure}
    
    \caption{\textbf{Qualitative segmentation results of \ours{}.} \textit{Closed-vocabulary segmentation (\cref{fig:voc_multi_qual_results} and \ref{fig:coco_qual_results}).} From top to bottom: original image, clustering of DM features, and final segmentation. Our pipeline filters out redundant clusters while retaining key objects and refines coarse masks to yield sharp segmentation maps. \textit{Open-vocabulary segmentation (\cref{fig:openvoc-qual})}}
    \label{fig:qualitatives}
\end{figure*}

\section{Discussion}

\subsubsection{Beyond Stable Diffusion.} 
While this work focuses on a Stable Diffusion, other DMs may also fit into our framework. 
Recent works~\cite{kang2023scalinggan} have shown that GAN models, when properly scaled, can perform on par with diffusion models. 
Exploring such recent large-scale generative models, including GANs and diffusion models~\cite{podell2023sdxl}, may lead to better segmentation performance.
\subsubsection{Beyond semantic segmentation.} This work is evaluated solely on semantic segmentation. However, we believe that the internal representations might be useful for other dense prediction tasks (\textit{e.g.}, Instance Segmentation, Depth Estimation, Object Detection), requiring rich localized and spatial representations.  
\subsubsection{Limitations.} 
This work has several limitations. 
First, the performance of \ours{} is still lagging behind SoTA approaches for segmentation, and not relying on any training or supervision will most likely hinder achieving SoTA results. 
Second, based on the captioner capability, the number of objects detected from the global description might not cover all the objects in the dataset. 
Finally, the work explores relatively simple academic benchmarks that do not necessarily reflect complex real-world cases where we have a large number of objects/classes to consider. 
\def\myim#1{\includegraphics[scale=0.2]{#1}}
\newcolumntype{M}[1]{>{\centering\arraybackslash}m{#1}}
\newcolumntype{T}[1]{>{\small\centering\arraybackslash}m{#1}}
\begin{figure}[t]
    \centering
    \begin{minipage}[t]{\linewidth}
        \centering
        \scalebox{0.45}{
            \begin{tabular}{M{16.5mm}M{28.2mm}M{28.2mm}M{28.2mm}M{28.2mm}}
                &\includegraphics[width=\linewidth]{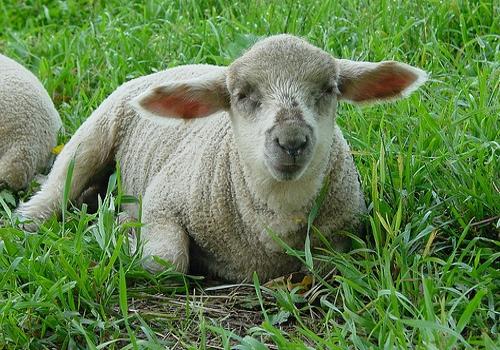} &
                \includegraphics[width=\linewidth]{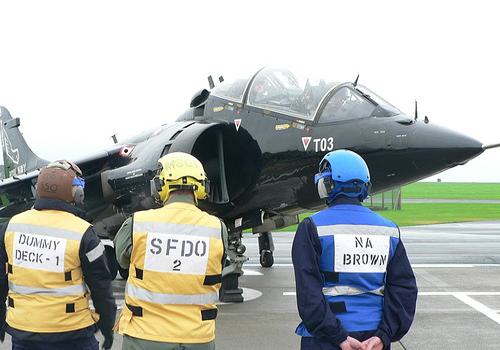} &
                \includegraphics[width=\linewidth]{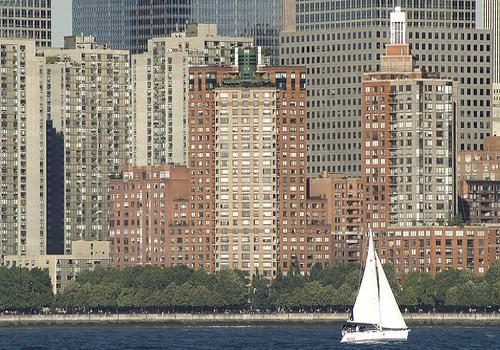} &
                \includegraphics[width=\linewidth]{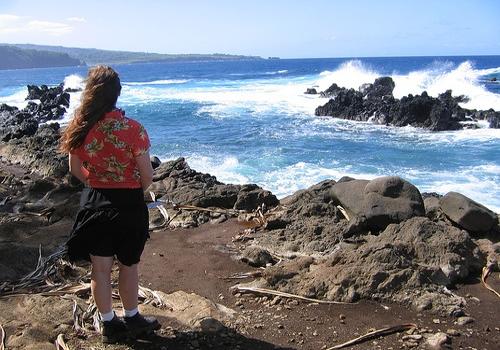}\\
                \multicolumn{1}{r}{\textbf{\rotatebox[origin=c]{90}{\small\textsc{SDv1.5}}}} &
                \includegraphics[width=\linewidth]{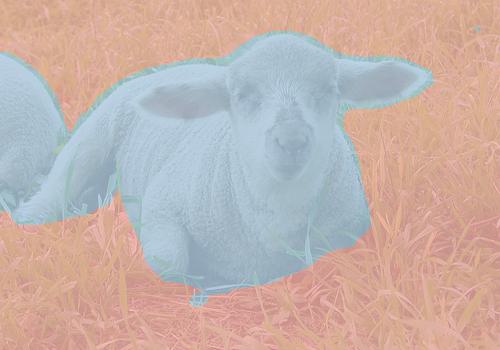} &
                \includegraphics[width=\linewidth]{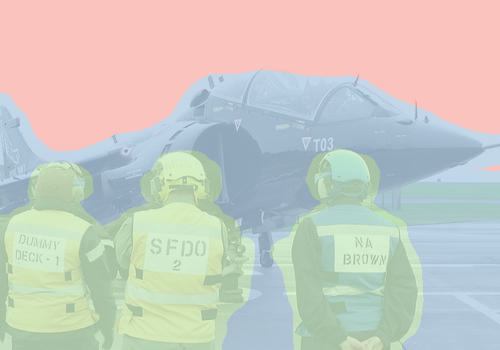} &
                \includegraphics[width=\linewidth]{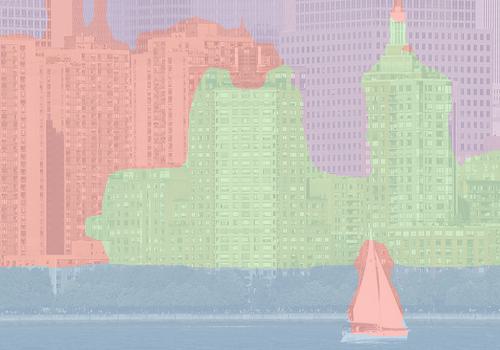} &
                \includegraphics[width=\linewidth]{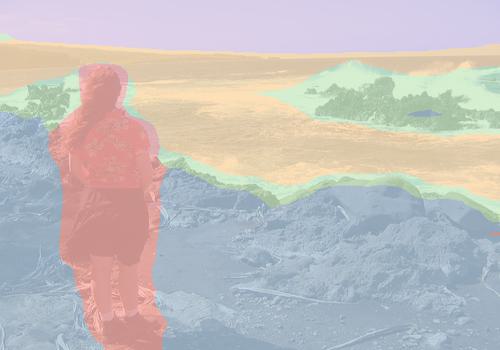} \\ 
                \multicolumn{1}{r}{\textbf{\rotatebox[origin=c]{90}{\small\textsc{VIT-H/14}}}} &
                \includegraphics[width=\linewidth]{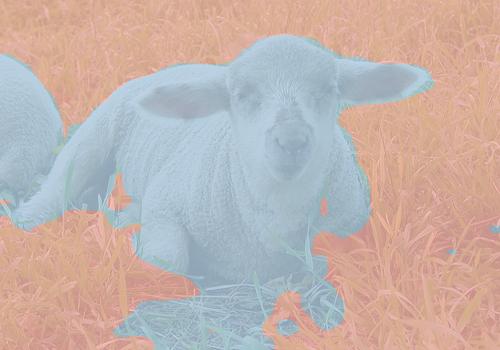} &
                \includegraphics[width=\linewidth]{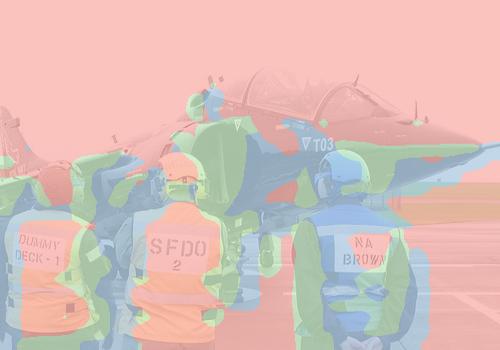} &
                \includegraphics[width=\linewidth]{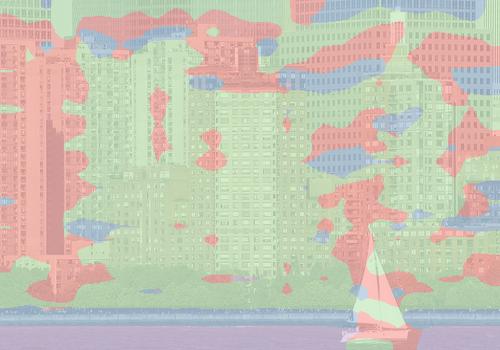} &
                \includegraphics[width=\linewidth]{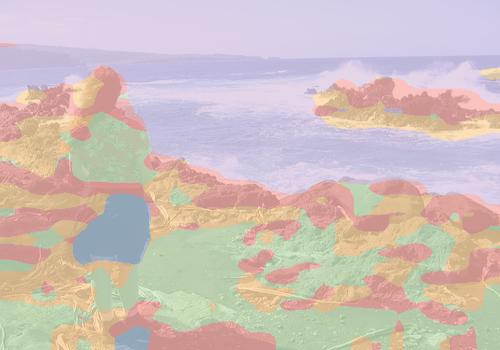} \\ 
                \multicolumn{1}{r}{\textbf{\rotatebox[origin=c]{90}{\small\textsc{DINOv2}}}} &
                \includegraphics[width=\linewidth]{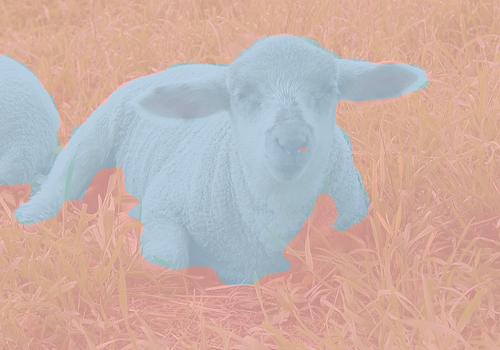} &
                \includegraphics[width=\linewidth]{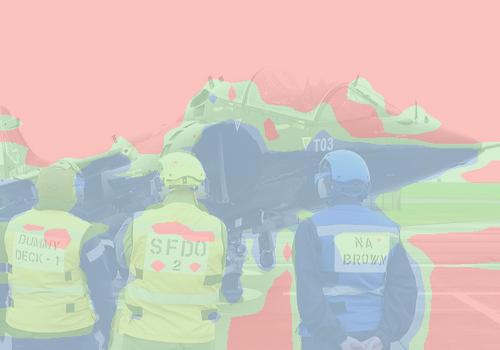} &
                \includegraphics[width=\linewidth]{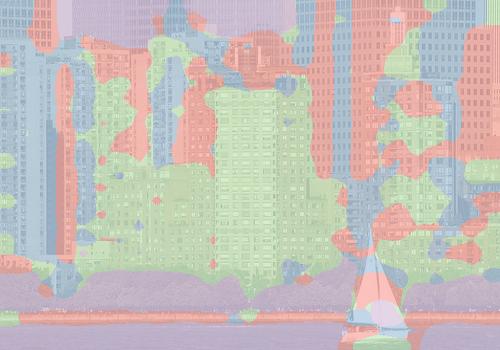} &
                \includegraphics[width=\linewidth]{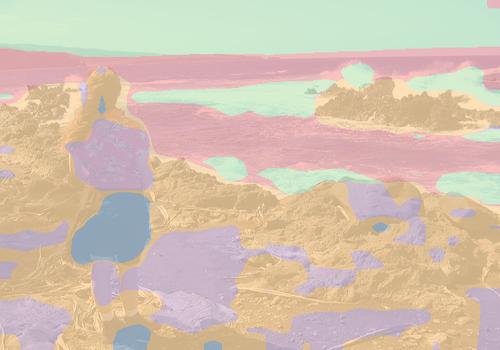} \\ 
                \multicolumn{1}{r}{\textbf{\rotatebox[origin=c]{90}{\small\textsc{\shortstack{CLIP \\ VIT-L/14}}}}} &
                \includegraphics[width=\linewidth]{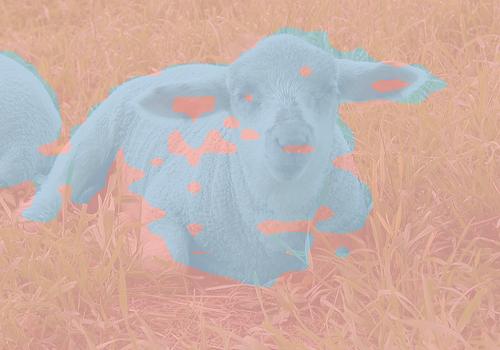} &
                \includegraphics[width=\linewidth]{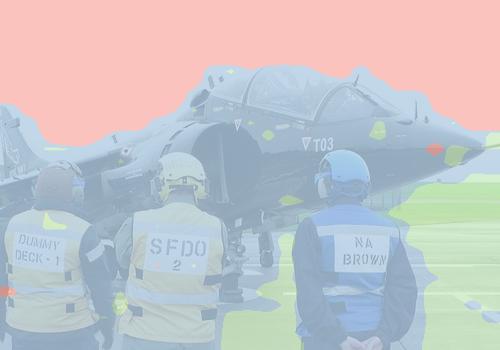} &
                \includegraphics[width=\linewidth]{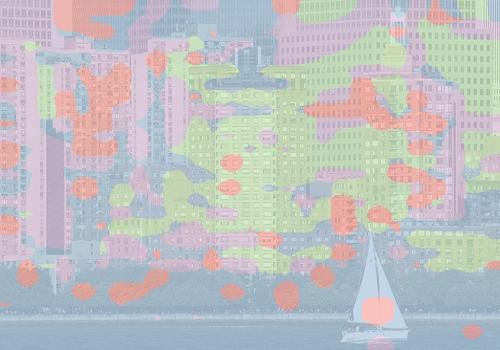} &
                \includegraphics[width=\linewidth]{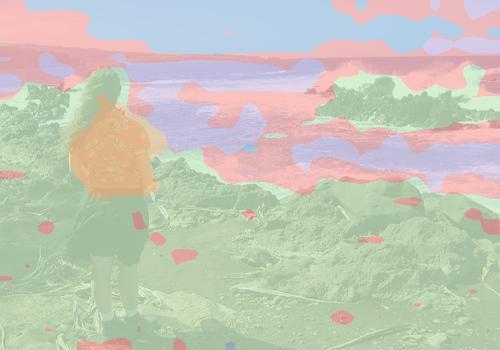} \\
                & $K=2$ & $K=3$ & $K=4$ & $K=5$
            \end{tabular}
            }

        \captionof{figure}{\textbf{Clustering features from different pre-trained backbones.} Stable diffusion model exhibits more semantic-aware internal features compared to those of CLIP-ViT, DINOv2 or VIT.}
        \label{fig:backbones_qual}
    \end{minipage}
\end{figure}

\begin{figure*}[t]
    \begin{minipage}[t]{0.32\linewidth}
    
        \captionof{table}{
            \textbf{Impact of each stage of \ours{}.} 
            We evaluate on Pascal VOC the baseline (SDM and CLIP, without caption support), then we add the captioning module (BLIP), the refinement module and the stable diffusion attention maps (Attn).
        }
        \label{tab:ablation}
        \centering
        \vspace{1em}
        \resizebox{0.8\linewidth}{!}{
            \begin{tabular}{lc}
                \hline
                \textbf{Method} & \textbf{mIoU} \\
                \hline
                Baseline & 17.52 \\
                BLIP & 50.39 \\
                BLIP+Refinement & 53.27 \\
                BLIP+Refinement+Attn & 53.31 \\
                \hline
            \end{tabular}
        }
        \begin{minipage}[t]{\linewidth}
            \centering
            \vspace{1em}
            \captionof{table}{
            \textbf{Comparison with other backbones.} 
            We evaluate on a subset of Pascal VOC the impact of clustering features from other backbones after resizing to the original image size. 
            Mask refinement is deactivated to better exhibit performance gaps.
            }
            \label{tab:backbones_quant}
            \resizebox{0.8\linewidth}{!}{
                \begin{tabular}{ccccc}
                    \hline
                    \textbf{Models} & 
                    \thead{\textbf{Pixel} \\ \textbf{Accuracy}} &
                    \thead{\textbf{mIoU}} \\
                    \hline
                    CLIP-ViT & 62.10 & 37.07 \\
                    ViT & 55.49 & 30.64 \\
                    DiNOv2 & 60.32 & 34.58 \\
                    \hline
                    SDM & 72.19 & 49.52 \\
                    \hline
                \end{tabular}
            }
        \end{minipage} 
    \end{minipage}
    \hfill
    \begin{minipage}[t]{0.32\textwidth}
        \centering
        \captionof{table}{
        \textbf{Impact of extracted features.}
        We evaluate \ours{} on Pascal VOC when clustering feature maps, attention maps and both feature and attention maps. We vary the resolution of the features and $K$ of clustering.
        Best results are highlighted \textbf{bold}, \textbf{\underline{underlined}}.} 
        \label{tab:ablation_layers}
        \resizebox{.9\linewidth}{!}{
            \begin{tabular}{cccc}
                \hline
                \thead{\textbf{Feature Maps} \\ \textbf{Resolution}} & \thead{$K$} & \thead{$\mathcal{F}$} & \thead{$\mathcal{A+F}$} \\ 
                \hline
                \multirow{3}{*}{16} &  $3$ & 52.07 & 51.81 \\ 
                ~ &  $4$ & \textbf{\underline{53.27}} & \textbf{53.31} \\ 
                ~ & $5$ & 51.85 & 52.04 \\ 
                \hline

                \multirow{3}{*}{32} &  $3$ & 46.73 & 46.06 \\ 
                ~ &  $4$ & 50.31 & 49.17 \\ 
                ~ & $5$ & 49.82 & 48.84 \\ 
                \hline

                \multirow{3}{*}{64} & $3$ & 42.54 & 40.15 \\ 
                ~ &  $4$ & 44.72 & 42.67 \\ 
                ~ & $5$ & 44.73 & 42.35 \\ 
                \hline

                \multirow{3}{*}{16+32} &  $3$ & 51.09 & 50.20 \\ 
                ~ & $4$ & 53.18 & 52.79 \\ 
                ~ & $5$ & 52.37 & 51.64 \\ 
                \hline

                \multirow{3}{*}{32+64} & $3$ & 46.35 & 45.07 \\ 
                ~ & $4$ & 49.90 & 47.99 \\ 
                ~ & $5$ & 50.03 & 48.20 \\ 
                \hline

                \multirow{3}{*}{16+32+64} & $3$ & 51.16 & 49.58 \\ 
                ~ & $4$ & 52.95 & 52.06 \\ 
                ~ & $5$ & 52.32 & 51.59 \\ 
                \hline
            \end{tabular}
        }
    \end{minipage}
    \hfill
    \begin{minipage}[t]{0.32\linewidth}
            \captionof{table}{
                \textbf{Effectiveness of refinement techniques.}
                We report the mIoU on Pascal VOC when different (or no) refinement techniques are used. Best in \textbf{bold} and \textbf{\underline{underlined}}.
            }
            \label{tab:ablation_refinement}
            \centering
            \resizebox{\linewidth}{!}{
                \begin{tabular}{ccccc}
                    \hline
                    \thead{\textbf{Feature maps} \\ \textbf{Resolution}} & \thead{\textbf{Refinement}} & \thead{$K$} & \thead{$\mathcal{F}$} 
                    & \thead{$\mathcal{A+F}$} \\ 
                    \hline
                    \multirow{9}{*}{16} & ~ & 3 & 48.95 &
                    48.66 \\ 
                    ~ & No Refinement & 4 & 50.39 &
                    50.77 \\ 
                    ~ & ~ & 5 & 49.49 &
                    50.03 \\ 
                    \cline{2-5}
    
                    ~ & ~ & 3 & 52.07 & 
                    51.81 \\ 
                    ~ & CRF & 4 & \underline{53.27} &
                    \textbf{53.31} \\ 
                    ~ & ~ & 5 & 51.85 &
                    52.04 \\ 
                    \cline{2-5}
    
                    ~ & ~ & 3 & 51.29 & 
                    51.01 \\ 
                    ~ & PAMR & 4 & 52.82 & 
                    52.66 \\ 
                    ~ & ~ & 5 & 51.79 &
                    51.99 \\ 
                    \hline
                \end{tabular}
            }
        \hfill
        \captionof{table}{
            \textbf{Effectiveness of clustering attention maps and/or feature maps.} We ablate adding the attention maps to the features from $16 \times 16$ layers before clustering. 
            CRF refinement module is active.  
            Best in \textbf{bold} and \textbf{\underline{underlined}}.
        }
        \vspace{1em}
        \label{tab:coco_coco_27_ablation_K}
        \resizebox{\linewidth}{!}{
            \begin{tabular}{ccccc}
                \hline
                \thead{\textbf{Feature Maps} \\ \textbf{Resolution}} & \thead{\textbf{Dataset}} & \thead{$K$} & \thead{$\mathcal{F}$} & \thead{$\mathcal{A+F}$} \\ 
                \hline
                \multirow{6}{*}{16} &
                \multirow{3}{*}{COCO-27} 
                  & $3$ & 31.06 & 30.75 \\ 
                ~ && $4$ & 34.03 & 33.92 \\ 
                ~ && $5$ & \textbf{\underline{34.94}} & \textbf{35.01} \\ 
                \cline{2-5}
                & \multirow{3}{*}{COCO} 
                  & $3$ & 28.26 & 27.97 \\ 
                ~ && $4$ & 31.01 & 30.94 \\ 
                ~ && $5$ & \textbf{\underline{31.66}} & \textbf{31.84} \\ 
                \hline
            \end{tabular}
        }

    \end{minipage}
\end{figure*}

\section{Ablation Study}
\label{sec:ablation}
\subsection{Which backbone to pick for zero-shot segmentation?}
Numerous approaches to perform semantic segmentation exploit the internal representations of pre-trained models. So, to help us choose the most suitable model, we begin by investigating the internal representations of recent powerful and widely used models from the following classes:
\begin{itemize}
    \item \textit{Fully-supervised model for image classification} (ViT~\cite{dosovitskiy2020imagevit}): a transformer trained to classify images, the loss is applied to the CLS token, which extracts global information about the image.
    \item \textit{Self-supervised model} (DINOv2~\cite{oquab2023dinov2}): a transformer trained without any human supervision on different objectives, such as forcing the same representations from different crops of the same image.
    \item \textit{Contrastive image-text aligned model} (CLIP~\cite{radford2021learning}): a transformer model consisting of both text and image encoders, trained to align the representation of the image features to the corresponding text features.
    \item \textit{Diffusion model} (Stable Diffusion (SD)~\cite{rombach2022high}): a text-to-image generative model, trained with a simple denoising objective.
\end{itemize}
We extract the features from the last or penultimate layer for each model and apply a K-means clustering with a fixed number of clusters.~\Cref{fig:backbones_qual} shows a qualitative comparison between these backbones. We can notice that the features extracted from the SD model show a clear superiority in terms of semantic alignment compared to other models. Motivated by such observation, we build our zero-shot segmentation pipeline on top of the SD model.

\subsection{Ablating different model components}
In this section, we ablate different components of \ours.~\Cref{tab:ablation} shows the effectiveness of each part. The baseline involves extraction of image features through the diffusion model, clustering of these features, and associating each generated mask with its corresponding class using CLIP. We notice that this baseline has a very low score on Pascal VOC, revealing that, despite having good localization capability, there is still a large gap to fill to obtain good segmentation scores.  It appears that BLIP is a key component as it is mainly used to reduce the number of candidate classes. In addition, refinement brings additional points. We notice that the additional use of attention maps does not bring significant improvement.

\subsection{Comparison with different backbones} 
To support the use of a diffusion U-Net as a feature extractor backbone for segmentation tasks, we conduct qualitative (\Cref{fig:backbones_qual}) and quantitative (\Cref{tab:backbones_quant}) experiments that highlight the intrinsic capacity of this architecture to extract semantic information. 
\Cref{tab:backbones_quant} shows the superiority of SDM compared to other backbones. SDM outperforms other backbones by large margins.
A potential rationale for this observation lies in the superior semantic information retention of the diffusion U-Net compared to alternative models, attributed to its inherent capacity to set a structural image layout, internally acquired during the training phase. 
Moreover, its architecture is closely related to a Feature Pyramid Network~\cite{lin2017feature} that has shown in the past significant improvement as a generic feature extractor in several dense prediction tasks. 
\subsection{The choice of $K$ in K-means} 
In this ablation study, we analyze the role of $K$ depending on the dataset. While $K=4$ yields the best performance in the case of Pascal VOC, as shown in~\Cref{tab:coco_coco_27_ablation_K}, COCO shows slightly better performance for $K=5$. This difference is likely due to the complexity of the latter dataset (\textit{i.e.}, more objects in the same image and very small objects). However, the results for $K=4$ and $K=5$ are comparable, so for computational reasons, we keep $K=4$.
\subsection{Which diffusion model features to use?} 
We investigate the importance of features extracted from different places in the diffusion model. In~\Cref{tab:ablation_layers}, we report the comparison between features extracted from blocks 16, 32 and 64. We also compare output features and both attention and features. We notice that using lower resolution features ($16$ or $16+32$) brings the best performance while picking features from single higher resolutions ($32$, $64$ and their combinations) does not improve the mIoU. For computational reasons (\textit{i.e.}, to minimize the cost of clustering while maintaining performance), we have chosen the best configuration in which only the feature maps are extracted from the SDM, and the number of clusters is fixed at $K=4$.
\subsection{Different refinement modules}
To improve the quality of semantic maps, we employed a refinement module. We propose the results for 2 different refinement algorithms: CRF~\cite{krahenbuhl2011efficient} and PAMR~\cite{araslanov2020single}  that do not rely on mask supervision.
~\Cref{tab:ablation_refinement} shows that both PAMR and CRF yield similar results, which are significantly better than the baseline (+3 points mIoU).
\section{Conclusion}
In this work, we propose a segmentation pipeline that leverages pre-trained diffusion models and does not rely on any additional training or supervision. The work uses several pre-trained models to extract good features, cluster them, and associate each cluster to a textual class. We show competitive results to existing approaches that are based on training or different kinds of supervision. We hope that this work will push for more efforts to focus on adapting existing foundation models for different existing and novel tasks, without rushing to large-scale training.

{
\bibliographystyle{IEEEtran}
\bibliography{main}

\begin{thebibliography}{10}
\providecommand{\url}[1]{#1}
\csname url@samestyle\endcsname
\providecommand{\newblock}{\relax}
\providecommand{\bibinfo}[2]{#2}
\providecommand{\BIBentrySTDinterwordspacing}{\spaceskip=0pt\relax}
\providecommand{\BIBentryALTinterwordstretchfactor}{4}
\providecommand{\BIBentryALTinterwordspacing}{\spaceskip=\fontdimen2\font plus
\BIBentryALTinterwordstretchfactor\fontdimen3\font minus \fontdimen4\font\relax}
\providecommand{\BIBforeignlanguage}[2]{{%
\expandafter\ifx\csname l@#1\endcsname\relax
\typeout{** WARNING: IEEEtran.bst: No hyphenation pattern has been}%
\typeout{** loaded for the language `#1'. Using the pattern for}%
\typeout{** the default language instead.}%
\else
\language=\csname l@#1\endcsname
\fi
#2}}
\providecommand{\BIBdecl}{\relax}
\BIBdecl

\bibitem{bommasani2021opportunities}
R.~Bommasani, D.~A. Hudson, E.~Adeli, R.~Altman, S.~Arora, S.~von Arx, M.~S. Bernstein, J.~Bohg, A.~Bosselut, E.~Brunskill \emph{et~al.}, ``On the opportunities and risks of foundation models,'' \emph{arXiv preprint arXiv:2108.07258}, 2021.

\bibitem{zhou2023comprehensive}
C.~Zhou, Q.~Li, C.~Li, J.~Yu, Y.~Liu, G.~Wang, K.~Zhang, C.~Ji, Q.~Yan, L.~He \emph{et~al.}, ``A comprehensive survey on pretrained foundation models: A history from bert to chatgpt,'' \emph{arXiv preprint arXiv:2302.09419}, 2023.

\bibitem{awais2023foundational}
M.~Awais, M.~Naseer, S.~Khan, R.~M. Anwer, H.~Cholakkal, M.~Shah, M.-H. Yang, and F.~S. Khan, ``Foundational models defining a new era in vision: A survey and outlook,'' \emph{arXiv preprint arXiv:2307.13721}, 2023.

\bibitem{achiam2023gpt}
J.~Achiam, S.~Adler, S.~Agarwal, L.~Ahmad, I.~Akkaya, F.~L. Aleman, D.~Almeida, J.~Altenschmidt, S.~Altman, S.~Anadkat \emph{et~al.}, ``Gpt-4 technical report,'' \emph{arXiv preprint arXiv:2303.08774}, 2023.

\bibitem{team2023gemini}
G.~Team, R.~Anil, S.~Borgeaud, Y.~Wu, J.-B. Alayrac, J.~Yu, R.~Soricut, J.~Schalkwyk, A.~M. Dai, A.~Hauth \emph{et~al.}, ``Gemini: a family of highly capable multimodal models,'' \emph{arXiv preprint arXiv:2312.11805}, 2023.

\bibitem{touvron2023llama}
H.~Touvron, T.~Lavril, G.~Izacard, X.~Martinet, M.-A. Lachaux, T.~Lacroix, B.~Rozi{\`e}re, N.~Goyal, E.~Hambro, F.~Azhar \emph{et~al.}, ``Llama: Open and efficient foundation language models,'' \emph{arXiv preprint arXiv:2302.13971}, 2023.

\bibitem{radford2021learning}
A.~Radford, J.~W. Kim, C.~Hallacy, A.~Ramesh, G.~Goh, S.~Agarwal, G.~Sastry, A.~Askell, P.~Mishkin, J.~Clark \emph{et~al.}, ``Learning transferable visual models from natural language supervision,'' in \emph{International conference on machine learning}.\hskip 1em plus 0.5em minus 0.4em\relax PMLR, 2021, pp. 8748--8763.

\bibitem{alayrac2022flamingo}
J.-B. Alayrac, J.~Donahue, P.~Luc, A.~Miech, I.~Barr, Y.~Hasson, K.~Lenc, A.~Mensch, K.~Millican, M.~Reynolds \emph{et~al.}, ``Flamingo: a visual language model for few-shot learning,'' \emph{Advances in Neural Information Processing Systems}, vol.~35, pp. 23\,716--23\,736, 2022.

\bibitem{shukor2023unified}
M.~Shukor, C.~Dancette, A.~Rame, and M.~Cord, ``Unified model for image, video, audio and language tasks,'' \emph{arXiv preprint arXiv:2307.16184}, 2023.

\bibitem{ramesh2022hierarchical}
A.~Ramesh, P.~Dhariwal, A.~Nichol, C.~Chu, and M.~Chen, ``Hierarchical text-conditional image generation with clip latents,'' \emph{arXiv preprint arXiv:2204.06125}, vol.~1, no.~2, p.~3, 2022.

\bibitem{brown2020language}
T.~Brown, B.~Mann, N.~Ryder, M.~Subbiah, J.~D. Kaplan, P.~Dhariwal, A.~Neelakantan, P.~Shyam, G.~Sastry, A.~Askell \emph{et~al.}, ``Language models are few-shot learners,'' \emph{Advances in neural information processing systems}, vol.~33, pp. 1877--1901, 2020.

\bibitem{kang2023scaling}
M.~Kang, J.-Y. Zhu, R.~Zhang, J.~Park, E.~Shechtman, S.~Paris, and T.~Park, ``Scaling up gans for text-to-image synthesis,'' in \emph{Proceedings of the IEEE/CVF Conference on Computer Vision and Pattern Recognition}, 2023, pp. 10\,124--10\,134.

\bibitem{zhang2023text}
C.~Zhang, C.~Zhang, M.~Zhang, and I.~S. Kweon, ``Text-to-image diffusion model in generative ai: A survey,'' \emph{arXiv preprint arXiv:2303.07909}, 2023.

\bibitem{denoisingdiffusionprobabilisticmodels}
\BIBentryALTinterwordspacing
J.~Ho, A.~Jain, and P.~Abbeel, ``Denoising diffusion probabilistic models,'' \emph{CoRR}, vol. abs/2006.11239, 2020. [Online]. Available: \url{https://arxiv.org/abs/2006.11239}
\BIBentrySTDinterwordspacing

\bibitem{denoisingdiffusionimplicitmodels}
J.~Song, C.~Meng, and S.~Ermon, ``Denoising diffusion implicit models,'' \emph{CoRR}, vol. abs/2010.02502, 2020.

\bibitem{rombach2022high}
R.~Rombach, A.~Blattmann, D.~Lorenz, P.~Esser, and B.~Ommer, ``High-resolution image synthesis with latent diffusion models,'' in \emph{Proceedings of the IEEE/CVF conference on computer vision and pattern recognition}, 2022, pp. 10\,684--10\,695.

\bibitem{zhang2023adding}
L.~Zhang, A.~Rao, and M.~Agrawala, ``Adding conditional control to text-to-image diffusion models,'' in \emph{Proceedings of the IEEE/CVF International Conference on Computer Vision}, 2023, pp. 3836--3847.

\bibitem{couairon2023zero}
G.~Couairon, M.~Careil, M.~Cord, S.~Lathuili{\`e}re, and J.~Verbeek, ``Zero-shot spatial layout conditioning for text-to-image diffusion models,'' in \emph{Proceedings of the IEEE/CVF International Conference on Computer Vision}, 2023, pp. 2174--2183.

\bibitem{podell2023sdxl}
D.~Podell, Z.~English, K.~Lacey, A.~Blattmann, T.~Dockhorn, J.~M{\"u}ller, J.~Penna, and R.~Rombach, ``Sdxl: improving latent diffusion models for high-resolution image synthesis,'' \emph{arXiv preprint arXiv:2307.01952}, 2023.

\bibitem{vallaeys2024improveddepalm}
T.~Vallaeys, M.~Shukor, M.~Cord, and J.~Verbeek, ``Improved baselines for data-efficient perceptual augmentation of llms,'' \emph{arXiv preprint arXiv:2403.13499}, 2024.

\bibitem{lialin2023scaling}
V.~Lialin, V.~Deshpande, and A.~Rumshisky, ``Scaling down to scale up: A guide to parameter-efficient fine-tuning,'' \emph{arXiv preprint arXiv:2303.15647}, 2023.

\bibitem{csurka2022semantic}
G.~Csurka, R.~Volpi, B.~Chidlovskii \emph{et~al.}, ``Semantic image segmentation: Two decades of research,'' \emph{Foundations and Trends{\textregistered} in Computer Graphics and Vision}, vol.~14, no. 1-2, pp. 1--162, 2022.

\bibitem{Rahman_2023_CVPR}
A.~Rahman, J.~M.~J. Valanarasu, I.~Hacihaliloglu, and V.~M. Patel, ``Ambiguous medical image segmentation using diffusion models,'' in \emph{Proceedings of the IEEE/CVF Conference on Computer Vision and Pattern Recognition (CVPR)}, June 2023, pp. 11\,536--11\,546.

\bibitem{tzelepi2021semantic}
M.~Tzelepi and A.~Tefas, ``Semantic scene segmentation for robotics applications,'' in \emph{2021 12th International Conference on Information, Intelligence, Systems \& Applications (IISA)}.\hskip 1em plus 0.5em minus 0.4em\relax IEEE, 2021, pp. 1--4.

\bibitem{pan2020cross}
B.~Pan, J.~Sun, H.~Y.~T. Leung, A.~Andonian, and B.~Zhou, ``Cross-view semantic segmentation for sensing surroundings,'' \emph{IEEE Robotics and Automation Letters}, vol.~5, no.~3, pp. 4867--4873, 2020.

\bibitem{chen2023edge}
C.~Chen, C.~Wang, B.~Liu, C.~He, L.~Cong, and S.~Wan, ``Edge intelligence empowered vehicle detection and image segmentation for autonomous vehicles,'' \emph{IEEE Transactions on Intelligent Transportation Systems}, 2023.

\bibitem{wang2022semi}
Y.~Wang, H.~Wang, Y.~Shen, J.~Fei, W.~Li, G.~Jin, L.~Wu, R.~Zhao, and X.~Le, ``Semi-supervised semantic segmentation using unreliable pseudo-labels,'' in \emph{Proceedings of the IEEE/CVF Conference on Computer Vision and Pattern Recognition}, 2022, pp. 4248--4257.

\bibitem{weakly}
Z.~Huang, X.~Wang, J.~Wang, W.~Liu, and J.~Wang, ``Weakly-supervised semantic segmentation network with deep seeded region growing,'' in \emph{2018 IEEE/CVF Conference on Computer Vision and Pattern Recognition}, 2018, pp. 7014--7023.

\bibitem{hwang2019segsort}
J.-J. Hwang, S.~X. Yu, J.~Shi, M.~D. Collins, T.-J. Yang, X.~Zhang, and L.-C. Chen, ``Segsort: Segmentation by discriminative sorting of segments,'' in \emph{Proceedings of the IEEE/CVF International Conference on Computer Vision}, 2019, pp. 7334--7344.

\bibitem{liang2023open}
F.~Liang, B.~Wu, X.~Dai, K.~Li, Y.~Zhao, H.~Zhang, P.~Zhang, P.~Vajda, and D.~Marculescu, ``Open-vocabulary semantic segmentation with mask-adapted clip,'' in \emph{Proceedings of the IEEE/CVF Conference on Computer Vision and Pattern Recognition}, 2023, pp. 7061--7070.

\bibitem{xu2022simple}
M.~Xu, Z.~Zhang, F.~Wei, Y.~Lin, Y.~Cao, H.~Hu, and X.~Bai, ``A simple baseline for open-vocabulary semantic segmentation with pre-trained vision-language model,'' in \emph{European Conference on Computer Vision}.\hskip 1em plus 0.5em minus 0.4em\relax Springer, 2022, pp. 736--753.

\bibitem{xu2023open}
J.~Xu, S.~Liu, A.~Vahdat, W.~Byeon, X.~Wang, and S.~De~Mello, ``Open-vocabulary panoptic segmentation with text-to-image diffusion models,'' in \emph{Proceedings of the IEEE/CVF Conference on Computer Vision and Pattern Recognition}, 2023, pp. 2955--2966.

\bibitem{baranchuk2021label}
D.~Baranchuk, I.~Rubachev, A.~Voynov, V.~Khrulkov, and A.~Babenko, ``Label-efficient semantic segmentation with diffusion models,'' \emph{International Conference on Learning Representations (ICLR)}, 2022.

\bibitem{wang2023imagen}
S.~Wang, C.~Saharia, C.~Montgomery, J.~Pont-Tuset, S.~Noy, S.~Pellegrini, Y.~Onoe, S.~Laszlo, D.~J. Fleet, R.~Soricut \emph{et~al.}, ``Imagen editor and editbench: Advancing and evaluating text-guided image inpainting,'' in \emph{Proceedings of the IEEE/CVF Conference on Computer Vision and Pattern Recognition}, 2023, pp. 18\,359--18\,369.

\bibitem{schuhmann2022laion5b}
C.~Schuhmann, R.~Beaumont, R.~Vencu, C.~Gordon, R.~Wightman, M.~Cherti, T.~Coombes, A.~Katta, C.~Mullis, M.~Wortsman, P.~Schramowski, S.~Kundurthy, K.~Crowson, L.~Schmidt, R.~Kaczmarczyk, and J.~Jitsev, ``Laion-5b: An open large-scale dataset for training next generation image-text models,'' 2022.

\bibitem{goodfellow2014generative}
I.~J. Goodfellow, J.~Pouget-Abadie, M.~Mirza, B.~Xu, D.~Warde-Farley, S.~Ozair, A.~Courville, and Y.~Bengio, ``Generative adversarial networks,'' 2014.

\bibitem{NEURIPS2019_0266e33d}
\BIBentryALTinterwordspacing
M.~Bucher, T.-H. VU, M.~Cord, and P.~P\'{e}rez, ``Zero-shot semantic segmentation,'' in \emph{Advances in Neural Information Processing Systems}, H.~Wallach, H.~Larochelle, A.~Beygelzimer, F.~d\textquotesingle Alch\'{e}-Buc, E.~Fox, and R.~Garnett, Eds., vol.~32.\hskip 1em plus 0.5em minus 0.4em\relax Curran Associates, Inc., 2019. [Online]. Available: \url{https://proceedings.neurips.cc/paper_files/paper/2019/file/0266e33d3f546cb5436a10798e657d97-Paper.pdf}
\BIBentrySTDinterwordspacing

\bibitem{dong2022maskclip}
X.~Dong, J.~Bao, Y.~Zheng, T.~Zhang, D.~Chen, H.~Yang, M.~Zeng, W.~Zhang, L.~Yuan, D.~Chen, F.~Wen, and N.~Yu, ``Maskclip: Masked self-distillation advances contrastive language-image pretraining,'' 2022.

\bibitem{Luddecke_2022}
\BIBentryALTinterwordspacing
T.~Luddecke and A.~Ecker, ``Image segmentation using text and image prompts,'' in \emph{2022 IEEE/CVF Conference on Computer Vision and Pattern Recognition (CVPR)}.\hskip 1em plus 0.5em minus 0.4em\relax IEEE, Jun. 2022. [Online]. Available: \url{http://dx.doi.org/10.1109/CVPR52688.2022.00695}
\BIBentrySTDinterwordspacing

\bibitem{yu2023convolutions}
Q.~Yu, J.~He, X.~Deng, X.~Shen, and L.-C. Chen, ``Convolutions die hard: Open-vocabulary segmentation with single frozen convolutional clip,'' 2023.

\bibitem{shin2022reco}
G.~Shin, W.~Xie, and S.~Albanie, ``Reco: Retrieve and co-segment for zero-shot transfer,'' \emph{Advances in Neural Information Processing Systems}, vol.~35, pp. 33\,754--33\,767, 2022.

\bibitem{imagenet}
J.~Deng, W.~Dong, R.~Socher, L.-J. Li, K.~Li, and L.~Fei-Fei, ``Imagenet: A large-scale hierarchical image database,'' in \emph{2009 IEEE Conference on Computer Vision and Pattern Recognition}, 2009, pp. 248--255.

\bibitem{xu2022groupvit}
J.~Xu, S.~De~Mello, S.~Liu, W.~Byeon, T.~Breuel, J.~Kautz, and X.~Wang, ``Groupvit: Semantic segmentation emerges from text supervision,'' in \emph{Proceedings of the IEEE/CVF Conference on Computer Vision and Pattern Recognition}, 2022, pp. 18\,134--18\,144.

\bibitem{ren2023viewco}
\BIBentryALTinterwordspacing
P.~Ren, C.~Li, H.~Xu, Y.~Zhu, G.~Wang, J.~Liu, X.~Chang, and X.~Liang, ``Viewco: Discovering text-supervised segmentation masks via multi-view semantic consistency,'' in \emph{The Eleventh International Conference on Learning Representations}, 2023. [Online]. Available: \url{https://openreview.net/forum?id=2XLRBjY46O6}
\BIBentrySTDinterwordspacing

\bibitem{xu2023learningovs}
J.~Xu, J.~Hou, Y.~Zhang, R.~Feng, Y.~Wang, Y.~Qiao, and W.~Xie, ``Learning open-vocabulary semantic segmentation models from natural language supervision,'' in \emph{Proceedings of the IEEE/CVF Conference on Computer Vision and Pattern Recognition}, 2023, pp. 2935--2944.

\bibitem{chen2023robustclassif}
H.~Chen, Y.~Dong, Z.~Wang, X.~Yang, C.~Duan, H.~Su, and J.~Zhu, ``Robust classification via a single diffusion model,'' \emph{arXiv preprint arXiv:2305.15241}, 2023.

\bibitem{mukhopadhyay2023diffusionclassif}
S.~Mukhopadhyay, M.~Gwilliam, V.~Agarwal, N.~Padmanabhan, A.~Swaminathan, S.~Hegde, T.~Zhou, and A.~Shrivastava, ``Diffusion models beat gans on image classification,'' \emph{arXiv preprint arXiv:2307.08702}, 2023.

\bibitem{chen2023diffusiondetect}
S.~Chen, P.~Sun, Y.~Song, and P.~Luo, ``Diffusiondet: Diffusion model for object detection,'' in \emph{Proceedings of the IEEE/CVF International Conference on Computer Vision}, 2023, pp. 19\,830--19\,843.

\bibitem{zhang2023diffusionenginedetect}
M.~Zhang, J.~Wu, Y.~Ren, M.~Li, J.~Qin, X.~Xiao, W.~Liu, R.~Wang, M.~Zheng, and A.~J. Ma, ``Diffusionengine: Diffusion model is scalable data engine for object detection,'' \emph{arXiv preprint arXiv:2309.03893}, 2023.

\bibitem{wu2023diffumask}
W.~Wu, Y.~Zhao, M.~Z. Shou, H.~Zhou, and C.~Shen, ``Diffumask: Synthesizing images with pixel-level annotations for semantic segmentation using diffusion models,'' 2023.

\bibitem{ma2023diffusionseg}
C.~Ma, Y.~Yang, C.~Ju, F.~Zhang, J.~Liu, Y.~Wang, Y.~Zhang, and Y.~Wang, ``Diffusionseg: Adapting diffusion towards unsupervised object discovery,'' 2023.

\bibitem{pnvr2023ldznet}
K.~Pnvr, B.~Singh, P.~Ghosh, B.~Siddiquie, and D.~Jacobs, ``Ld-znet: A latent diffusion approach for text-based image segmentation,'' 2023.

\bibitem{karazija2023ovdiff}
L.~Karazija, I.~Laina, A.~Vedaldi, and C.~Rupprecht, ``Diffusion models for zero-shot open-vocabulary segmentation,'' \emph{arXiv preprint arXiv:2306.09316}, 2023.

\bibitem{burgert2022peekaboo}
R.~Burgert, K.~Ranasinghe, X.~Li, and M.~S. Ryoo, ``Peekaboo: Text to image diffusion models are zero-shot segmentors,'' \emph{arXiv preprint arXiv:2211.13224}, 2022.

\bibitem{li2023blip2}
J.~Li, D.~Li, S.~Savarese, and S.~Hoi, ``Blip-2: Bootstrapping language-image pre-training with frozen image encoders and large language models,'' 2023.

\bibitem{krahenbuhl2011efficient}
P.~Kr{\"a}henb{\"u}hl and V.~Koltun, ``Efficient inference in fully connected crfs with gaussian edge potentials,'' \emph{Advances in neural information processing systems}, vol.~24, 2011.

\bibitem{Everingham15}
M.~Everingham, S.~M.~A. Eslami, L.~Van~Gool, C.~K.~I. Williams, J.~Winn, and A.~Zisserman, ``The pascal visual object classes challenge: A retrospective,'' \emph{International Journal of Computer Vision}, vol. 111, no.~1, pp. 98--136, Jan. 2015.

\bibitem{lin2014microsoft}
T.-Y. Lin, M.~Maire, S.~Belongie, L.~Bourdev, R.~Girshick, J.~Hays, P.~Perona, D.~Ramanan, C.~L. Zitnick, and P.~Dollár, ``Microsoft coco: Common objects in context,'' 2014.

\bibitem{ranasinghe2022perceptual}
K.~Ranasinghe, B.~McKinzie, S.~Ravi, Y.~Yang, A.~Toshev, and J.~Shlens, ``Perceptual grouping in vision-language models,'' \emph{arXiv preprint arXiv:2210.09996}, 2022.

\bibitem{cha2023learningtcl}
J.~Cha, J.~Mun, and B.~Roh, ``Learning to generate text-grounded mask for open-world semantic segmentation from only image-text pairs,'' in \emph{Proceedings of the IEEE/CVF Conference on Computer Vision and Pattern Recognition}, 2023, pp. 11\,165--11\,174.

\bibitem{touvron2021trainingdeit}
H.~Touvron, M.~Cord, M.~Douze, F.~Massa, A.~Sablayrolles, and H.~J{\'e}gou, ``Training data-efficient image transformers \& distillation through attention,'' in \emph{International conference on machine learning}.\hskip 1em plus 0.5em minus 0.4em\relax PMLR, 2021, pp. 10\,347--10\,357.

\bibitem{caron2021emergingdino}
M.~Caron, H.~Touvron, I.~Misra, H.~J{\'e}gou, J.~Mairal, P.~Bojanowski, and A.~Joulin, ``Emerging properties in self-supervised vision transformers,'' in \emph{Proceedings of the IEEE/CVF international conference on computer vision}, 2021, pp. 9650--9660.

\bibitem{he2020momentummoco}
K.~He, H.~Fan, Y.~Wu, S.~Xie, and R.~Girshick, ``Momentum contrast for unsupervised visual representation learning,'' in \emph{Proceedings of the IEEE/CVF conference on computer vision and pattern recognition}, 2020, pp. 9729--9738.

\bibitem{zhou2022extractmaskclip}
C.~Zhou, C.~C. Loy, and B.~Dai, ``Extract free dense labels from clip,'' in \emph{European Conference on Computer Vision}.\hskip 1em plus 0.5em minus 0.4em\relax Springer, 2022, pp. 696--712.

\bibitem{align}
C.~Jia, Y.~Yang, Y.~Xia, Y.-T. Chen, Z.~Parekh, H.~Pham, Q.~Le, Y.-H. Sung, Z.~Li, and T.~Duerig, ``Scaling up visual and vision-language representation learning with noisy text supervision,'' in \emph{International conference on machine learning}.\hskip 1em plus 0.5em minus 0.4em\relax PMLR, 2021, pp. 4904--4916.

\bibitem{kang2023scalinggan}
M.~Kang, J.-Y. Zhu, R.~Zhang, J.~Park, E.~Shechtman, S.~Paris, and T.~Park, ``Scaling up gans for text-to-image synthesis,'' in \emph{Proceedings of the IEEE/CVF Conference on Computer Vision and Pattern Recognition}, 2023, pp. 10\,124--10\,134.

\bibitem{dosovitskiy2020imagevit}
A.~Dosovitskiy, L.~Beyer, A.~Kolesnikov, D.~Weissenborn, X.~Zhai, T.~Unterthiner, M.~Dehghani, M.~Minderer, G.~Heigold, S.~Gelly \emph{et~al.}, ``An image is worth 16x16 words: Transformers for image recognition at scale,'' \emph{arXiv preprint arXiv:2010.11929}, 2020.

\bibitem{oquab2023dinov2}
M.~Oquab, T.~Darcet, T.~Moutakanni, H.~Vo, M.~Szafraniec, V.~Khalidov, P.~Fernandez, D.~Haziza, F.~Massa, A.~El-Nouby \emph{et~al.}, ``Dinov2: Learning robust visual features without supervision,'' \emph{arXiv preprint arXiv:2304.07193}, 2023.

\bibitem{lin2017feature}
T.-Y. Lin, P.~Dollár, R.~Girshick, K.~He, B.~Hariharan, and S.~Belongie, ``Feature pyramid networks for object detection,'' 2017.

\bibitem{araslanov2020single}
N.~Araslanov and S.~Roth, ``Single-stage semantic segmentation from image labels,'' in \emph{Proceedings of the IEEE/CVF conference on computer vision and pattern recognition}, 2020, pp. 4253--4262.

\end{thebibliography}
}
\end{document}